\documentclass[a4paper]{article}

\usepackage[english]{babel} 
\usepackage{tikz}
\usepackage{graphicx}
\usetikzlibrary{calc,shapes.multipart,chains,arrows,fit,shapes,calc,backgrounds,trees}
\usepackage{url}
\usepackage[utf8]{inputenc}
\usepackage{epsfig}
\usepackage{calc}
\usepackage{amssymb}
\usepackage{amstext}
\usepackage{amsmath}
\usepackage{amsthm}
\usepackage[FIGTOPCAP,normal]{subfigure}
\usepackage{booktabs}

\title{HEPGAME and the Simplification of Expressions}
\author{Ben Ruijl$^{1,2}$, Jos Vermaseren$^1$, Aske Plaat$^2$, Jaap van den Herik$^2$\\$^1~$Nikhef, Science Park 105, 1098 XG Amsterdam, The Netherlands\\$^2~$Leiden University, Niels Bohrweg 1, 2333 CA Leiden, The Netherlands}

\date{}
\begin{document}
\maketitle
\thispagestyle{empty}

\begin{abstract}
Advances in high energy physics have created the need to increase computational capacity. Project HEPGAME was composed to address this challenge. One of the issues is that numerical integration of expressions of current interest have millions of terms and takes weeks to compute. We have investigated ways to simplify these expressions, using Horner schemes and common subexpression elimination. Our approach applies MCTS, a search procedure that has been successful in AI. We use it to find near-optimal Horner schemes. Although MCTS finds better solutions, this approach gives rise to two further challenges. (1) MCTS (with UCT) introduces a constant, $C_p$ that governs the balance between exploration and exploitation. This constant has to be tuned manually. (2) There should be more guided exploration at the bottom of the tree, since the current approach reduces the quality of the solution towards the end of the expression. We investigate NMCS (Nested Monte Carlo Search) to address both issues, but find that NMCS is computationally unfeasible for our problem. Then, we modify the MCTS formula by introducing a dynamic exploration-exploitation parameter $T$ that decreases linearly with the iteration number. Consequently, we provide a performance analysis. We observe that a variable $C_p$ solves our domain: it yields more exploration at the bottom and as a result the tuning problem has been simplified. The region in $C_p$ for which good values are found is increased by more than a tenfold. This result encourages us to continue our research to solve other prominent problems in High Energy Physics.
\end{abstract}

\section{Introduction}
\label{sec:intro}

High energy physics is a field that pushes the boundaries of possibilities, mathematically, technologically, and computationally. In String Theory new mathematics has been developed to describe the fundamentals of the universe \cite{Aspinwall2009}. At CERN and other institutes such as Fermilab, high accuracy particle detectors measure newly created particles from scattering processes (see, e.g., \cite{Aad2012}). And computationally, results from calculations of the Standard Model of particle physics and next generation theories, such as supersymmetry, are being compared to observational data to further our knowledge about the fundamental structure of nature (see, e.g, \cite{Ellis1989}).

Computational problems arise due to the fact that the expressions derived from the mentioned models contain hard integrals and that their size grows exponentially as more precision is required. As a result, for some processes the detectors at CERN are able to measure higher precision than the results obtained from theory. The project HEPGAME (``High Energy Physics Game'') aims to improve the current level of computational accuracy by applying methods from artificial intelligence to solve difficult problems \cite{hepgame}. 

Our work is based on FORM \cite{Kuipers2012,Kuipers2013B}. FORM is a computer algebra system for highly efficient manipulations of very large expressions. It contains special routines for tensors, gamma matrices, and other physics related objects. Our results will be part of the next release of FORM.

One of the current problems is the simplification of expressions. From Quantum Field Theory, expressions with millions of terms, taking up gigabytes of memory in intermediate form, are not uncommon. These expressions have to be numerically integrated, in order to match the theoretical results with the observations performed. For these large expressions, the integration process could take months. If we are able to reduce the number of operations, we are able to make high accuracy predictions that were previously computationally unfeasible.

In this paper we give an overview of the recent results on simplifying expressions. First, we describe two methods which we use to reduce the number of operations, namely Horner's rule for multivariate polynomials, and common subexpression elimination. Horner's rule is extracting variables outside brackets \cite{Horner1819}. For multivariate expressions the order of these variables is called a \emph{Horner scheme}. Next, we remove the common subexpressions \cite{Knuth1997}. The problem of finding the order that yields the least number of operations is NP-hard \cite{Ceberio2004}. We will investigate three methods of finding a near-optimal Horner scheme.

The first is Monte Carlo Tree Search (MCTS), using UCT as best-child criterion. We obtain simplifications that are up to $24$ times smaller for our benchmark polynomials \cite{Kuipers2013}. However, UCT is not straightforward, as it introduces an exploration-exploitation constant $C_p$ that must be tuned. Furthermore, it does little exploration at the bottom of the tree.

To address both issues, the second method investigated is Nested Monte Carlo Search (NMCS). NMCS does not not have the two challenges mentioned. However, our evaluation function turns out to be quite expensive (6 seconds for one of our benchmark polynomials). So, NMCS performs (too) many evaluations to find a path in the tree, rendering it unsuitable for our simplification task.

Third, we make a modification to UCT, which we call SA-UCT (Simulated Annealing UCT). SA-UCT introduces a dynamic exploration-exploitation parameter $T(i)$ that decreases linearly with the iteration number $i$. SA-UCT causes a gradual shift from exploration at the start of the simulation to exploitation at the end. As a consequence, the final iterations will be used for exploitation, improving their solution quality. Additionally, more branches reach the final states, resulting in more exploration at the bottom of the tree. Moreover, we show that the tuning of $C_p$ has become easier, since the region with appropriate values for $C_p$ has increased by at least a tenfold \cite{Ruijl2014}. The main contribution of SA-UCT is that this simplification of tuning allows for the results of our MCTS approach to be obtained much faster. In turn, our process is able to reduce the computation times of numerical integration from weeks to days or even hours.

This paper is structured as follows. Section \ref{sec:related} shows related work, section \ref{sec:background} provides a background on the optimization methods, section \ref{sec:mcts} introduces MCTS, section \ref{sec:nmcs} discusses NMCS, section \ref{sec:sa-uct} describes SA-UCT, section \ref{sec:conclusion} provides our conclusion and section \ref{sec:discussion} contains a discussion and an outlook for project HEPGAME.

\section{Related work}
\label{sec:related}

Computer algebra systems, expression simplification, and boolean problems are closely related. General purpose packages such as Mathematica and Maple have evolved out of early systems created by physicists and artificial intelligence in the 1960s. A prime example for the first development was the work by the later Nobel Prize laureate Martinus Veltman, who designed a program for symbolic mathematics, especially High Energy Physics, called Schoonschip (Dutch for ``clean ship,'' or ``clean up'') in 1963. In the course of the 1960s Gerard 't Hoofd accompanied Veltman with whom he shared the Nobel Prize. The first popular computer algebra systems were muMATH, Reduce, Derive (based on muMATH), and Macsyma. It is interesting to note that FORM \cite{Kuipers2012}, the system that we use in our work, is a direct successor to Schoonschip. 

The Boolean Satisfiability problem, or SAT, is a central problem in symbolic logic and computational complexity. Ever since Cook's seminal work \cite{Cook1971}, finding efficient solvers for SAT has driven much progress in computational logic and combinatorial optimization. MCTS has been quite successful in adversary search and optimization \cite{Browne2012}. In the current work, we discuss the application of MCTS to expression simplification. Curiously, we are not aware of many other works, with the notable exception of Previti et al. \cite{Previti2011}.

Expression simplification is a widely studied problem. We have already mentioned Horner schemes \cite{Knuth1997}, and common subexpression elimination (CSEE) \cite{dragon}, but there are several other methods, such as partial syntactic factorization \cite{Leiserson2010} and Breuer's growth algorithm \cite{Breuer1969}. Horner schemes and CSEE do not require much algebra: only the commutative and associative properties of the operators are used. Much research is put into simplifications using more algebraic properties, such as factorization, especially because of its interest for cryptographic research. 

In section \ref{sec:sa-uct} we will introduce modifications to UCT, in order to make the importance of exploration versus exploitation iteration-number dependent. In the past related changes have been proposed. For example, Discounted UCB \cite{Kocsis2006B} and Accelerated UCT \cite{Hashimoto2012} both modify the average score of a node to discount old wins over new ones. The difference between our method and past work is that the previous modifications alter the importance of exploring based on the history and do not guarantee that the focus shifts from exploration to exploitation. In contrast, this work focuses on the exploration-exploitation constant $C_p$ and on the role of exploration during the simulation.

\section{Background}
\label{sec:background}
In this section we discuss the two methods of reducing the number of operations. The first is Horner schemes, and the second is common subexpression elimination.

\subsection{Horner Schemes}
Horner's rule reduces the number of multiplications in an expression by lifting variables outside brackets \cite{Horner1819,Knuth1997,Ceberio2004}. For multivariate expressions Horner's rule can be applied sequentially, once for each variable. The order of this sequence is called the \emph{Horner scheme}. Take for example:
\begin{equation}
x^2z + x^3y + x^3yz \rightarrow x^2(z + x(y(1 + z))
\end{equation}
Here, first the variable $x$ is extracted (i.e., $x^2$ and $x$) and second, $y$. The number of multiplications is now reduced from $8$ to $4$. However, the order $x, y$ is chosen arbitrarily. One could also try the order $y, x$:
\begin{equation}
x^2z + x^3y + x^3yz \rightarrow x^2z + y(x^3(1 + z))
\end{equation}
for which the number of multiplications is $6$. Evidently, this is a suboptimal Horner scheme. There are $n!$ orders of extracting variables, where $n$ is the number of variables, and it turns out that the problem of selecting an optimal ordering is NP-hard \cite{Ceberio2004}. 

A heuristic that works reasonably well is selecting variables according to how frequently a term with such a variable occurs (``occurrence order''). A counter-example that shows that this is not always optimal is
\begin{equation}
x^{50}y + x^{40} + y + yz
\end{equation}
where extracting the most occurring variable $y$ first causes the $x^{50}$ and $x^{40}$ to end up in different subparts of the polynomial, preventing their common terms from being extracted. We note that ordering the variables according to its highest power or to the sum of its powers in all the terms, leads to other counter-examples.

Since the only property that Horner's rule requires is that the operator is associative, it follows that this method could also be used to reduce the number of operations in boolean expressions with logical operators. 

\subsection{Common Subexpression Elimination}
\begin{figure}[h]
\centering
\includegraphics[scale=1]{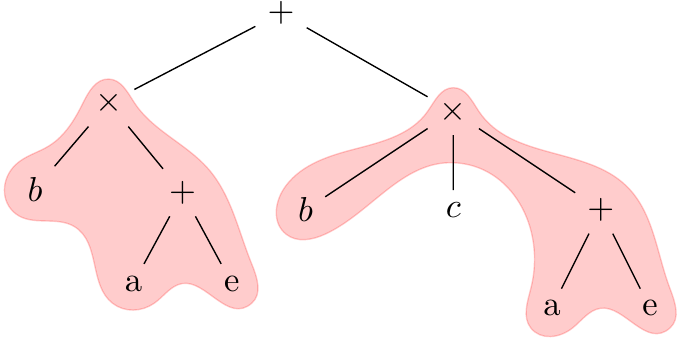}
\caption{A common subexpression (shaded) in an associative and commutative tree representation.}
\label{fig:cse}
\end{figure}

The number of operations can be reduced further by applying common subexpression elimination (CSEE). This method is well known from the fields of compiler construction \cite{dragon} and computer chess \cite{Adelson1988}, where it is applied to much smaller expressions or subtrees than in high energy physics. 
Figure \ref{fig:cse} shows an example of a common subexpression in a tree representation of an expression. The shaded expression $b(a+e)$ appears twice, and its removal means removing one superfluous addition and one multiplication.

CSEE is able to reduce both the number of multiplications and the number of additions, whereas Horner schemes are only able to reduce the number of multiplications.

We note that there is an interplay between Horner's rule and CSEE: a particularly good Horner scheme may reduce the number of multiplications the most, but the resulting scheme may expose less common subexpressions than a mediocre Horner scheme. Thus, we need a way to find a Horner scheme that reduces the number of operations the most after both Horner and CSEE have been applied. To achieve this situation, we investigate Monte Carlo Tree Search.

\section{Monte Carlo Tree Search}
\label{sec:mcts}

Monte Carlo Tree Search (MCTS) is a tree search method that has been successful in games such as Go, Hex, and other applications with a large state space \cite{Lee2009, Hayward2009}. It works by selectively building a tree, expanding only branches it deems worthwhile to explore. MCTS consists of four steps, which are displayed in figure \ref{fig:mcts}. The first step (\ref{fig:mcts}a) is the selection step, where a leaf or a not fully expanded node is selected according to some criterion (see below). Our choice is node $z$. In the expansion step (\ref{fig:mcts}b), a random unexplored child of the selected node is added to the tree (node $y$). In the simulation step (\ref{fig:mcts}(c)), the rest of the path to a final node is completed using random child selection. Finally a score $\Delta$ is obtained that signifies the score of the chosen path through the state space. In the backprogagation step (\ref{fig:mcts}(d)), this value is propagated back through the tree, which affects the average score (winrate) of a node (see below). The tree is built iteratively by repeating the four steps.

In the game of Go, each node represents a player move and in the expansion phase the game is played out, in basic implementations, by random moves. In the best performing implementations heuristics and pattern knowledge are used to complement a random playout \cite{Lee2009}. The final score is 1 if the game is won, and 0 if the game is lost. The entire tree is built, ultimately, to select the best first move. 

For our purposes, we need to build a complete Horner scheme, variable by variable. As such, each node will represent a variable and the depth of a node in the tree represents the position in the Horner scheme. Thus, in figure \ref{fig:mcts}(c) the partial Horner scheme is x,z,y and the rest of the scheme is filled in randomly with unused variables. The score of a path in our case, is the improvement of the path on the number of operations: the original number of operations divided by the number of operations after the Horner scheme and CSEE have been applied. We note that for our purposes the \emph{entire} Horner scheme is important and not just the first variable.

\tikzset{
  treenode/.style = {align=center, inner sep=0pt, text centered,
    font=\sffamily},
  arn_r/.style = {treenode, circle, black, draw=black, text width=1.5em,semithick,-}, 
  ar/.style = {arn_r,->,>=stealth,very thick},
  arr/.style = {arn_r,<-,>=stealth,very thick}
}

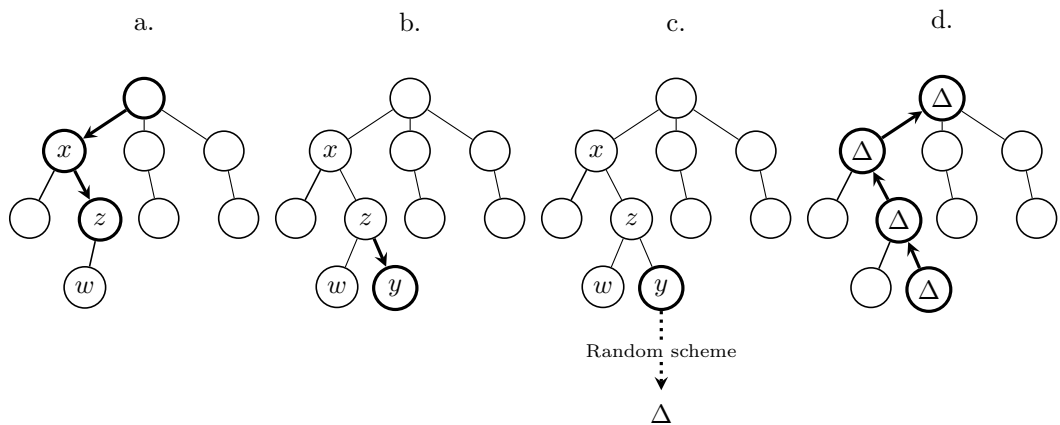
\begin{figure*}[hbt!]
\centering
\begin{tikzpicture}[level/.style={sibling distance = 1.5cm/#1,
  level distance = 1.0cm}] 
  
\begin{scope}[scale=0.7]
\node [ar] (r1) {}
    child[ar]{ node [ar] {$x$} 
            child[arn_r,below left]{ node [arn_r] {}
            }
            child[ar,below right]{ node [ar] {$z$}
							child[arn_r,below left]{ node [arn_r] {$w$}}
            }                            
    }
    child{ node [arn_r] {} 
    	child[below right] { node [arn_r] {} 
    	} 
    }
    child{ node [arn_r] {}
            child[below right]{ node [arn_r] {}
            }
		}
;
\node[above= 0.5cm of r1] {a.};
\end{scope}

\begin{scope}[scale=0.7,shift={(5,0)}]
\node [arn_r] (r2) {}
    child[]{ node [arn_r] {$x$} 
            child[arn_r,below left]{ node [arn_r] {} 
            }
            child[below right]{ node [arn_r] {$z$}
							child[below left]{ node [arn_r] {$w$}}
							child[ar,below right]{ node [ar] {$y$} }
            }                            
    }
    child{ node [arn_r] {} 
    	child[below right] { node [arn_r] {} 
    	} 
    }
    child{ node [arn_r] {}
            child[below right]{ node [arn_r] {}
            }
		}
;
\node[above= 0.5cm of r2] {b.};
\end{scope}

\begin{scope}[scale=0.7,shift={(10,0)}]
\node [arn_r] (r3) {}
    child[]{ node [arn_r] {$x$} 
            child[arn_r,below left]{ node [arn_r] {} 
            }
            child[below right]{ node [arn_r] {$z$}
							child[below left]{ node [arn_r] {$w$}}
							child[below right]{ node[ar] (ex) {$y$} }
            }                            
    }
    child{ node [arn_r] {} 
    	child[below right] { node [arn_r] {} 
    	} 
    }
    child{ node [arn_r] {}
            child[below right]{ node [arn_r] {}
            }
		}
; 
\node[above= 0.5cm of r3] {c.};
\draw[dotted,ar] (ex.south) to node[rectangle,anchor=center, text width=2cm,midway,font=\scriptsize,fill=white] {\text{Random scheme}} ++(0,-1.5cm) node[below] {$\Delta$};
\end{scope}

\begin{scope}[scale=0.7,shift={(15,0)}]
\node [arr] (r4) {$\Delta$}
    child[arr]{ node [arr] {$\Delta$} 
            child[arn_r,below left]{ node [arn_r] {} 
            }
            child[arr,below right]{ node [arr] {$\Delta$}
							child[arn_r,below left]{ node [arn_r] {}}
							child[arr,below right]{ node[arr] (ex) {$\Delta$} }
            }                            
    }
    child{ node [arn_r] {} 
    	child[below right] { node [arn_r] {} 
    	} 
    }
    child{ node [arn_r] {}
            child[below right]{ node [arn_r] {}
            }
		}
; 
\node[above= 0.5cm of r4] {d.};
\end{scope}

\end{tikzpicture}
\caption{An overview of the four phases of MCTS: selection (a), expansion (b), simulation (c), and backpropagation (d). The selection of a not fully expanded node is done using the best child criterion (in our case UCT, see below). $\Delta$ is the number of operations left in the final expression, after the Horner scheme and CSEE have been applied. See also \cite{Browne2012}.}
\label{fig:mcts}
\end{figure*}

In many MCTS implementations UCT (formula \eqref{eq:uct}) is chosen as the selection criterion \cite{Browne2012, Kocsis2006}:
\begin{equation}
\underset{\text{children $c$ of $s$}}{\operatorname{argmax}} \bar{x}(c) + 2 C_p \sqrt{\frac{2 \ln n(s)}{n(c)}}
\label{eq:uct}
\end{equation}
where $c$ is a child node of node $s$, $\bar{x}(c)$ the average score of node $c$, $n(c)$ the number of times the node $c$ has been visited, $C_p$ the exploration-exploitation constant, and $\operatorname{argmax}$ the function that selects the child with the maximum value. This formula balances exploitation, i.e., picking terms with a high average score, and exploration, i.e., selecting nodes where the child has not been visited often compared to the parent. The $C_p$ constant determines how strong the exploration term is: for high $C_p$ the focus will be on exploration, and for low $C_p$ the focus will be on exploitation.

\subsection{\label{sec:sens}Sensitivity Analysis}

\begin{figure}
\centering
\subfigure[$N=300$]{ \includegraphics[scale=0.572]{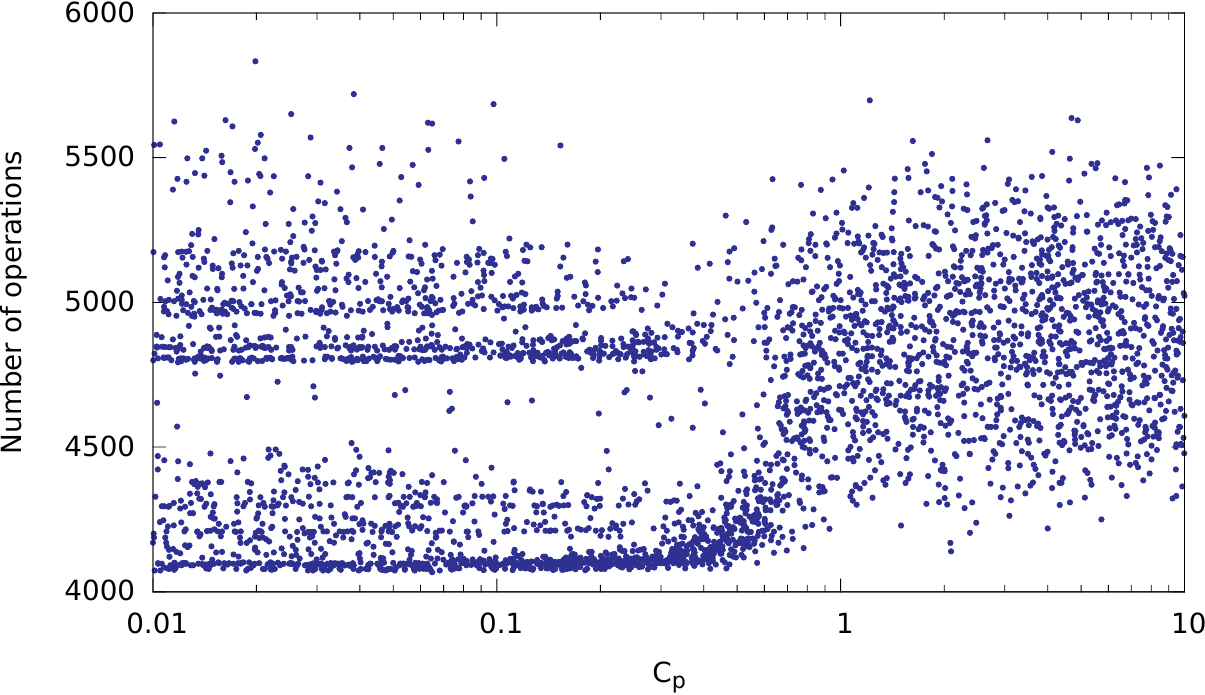} }
\subfigure[$N=3000$]{ \includegraphics[scale=0.572]{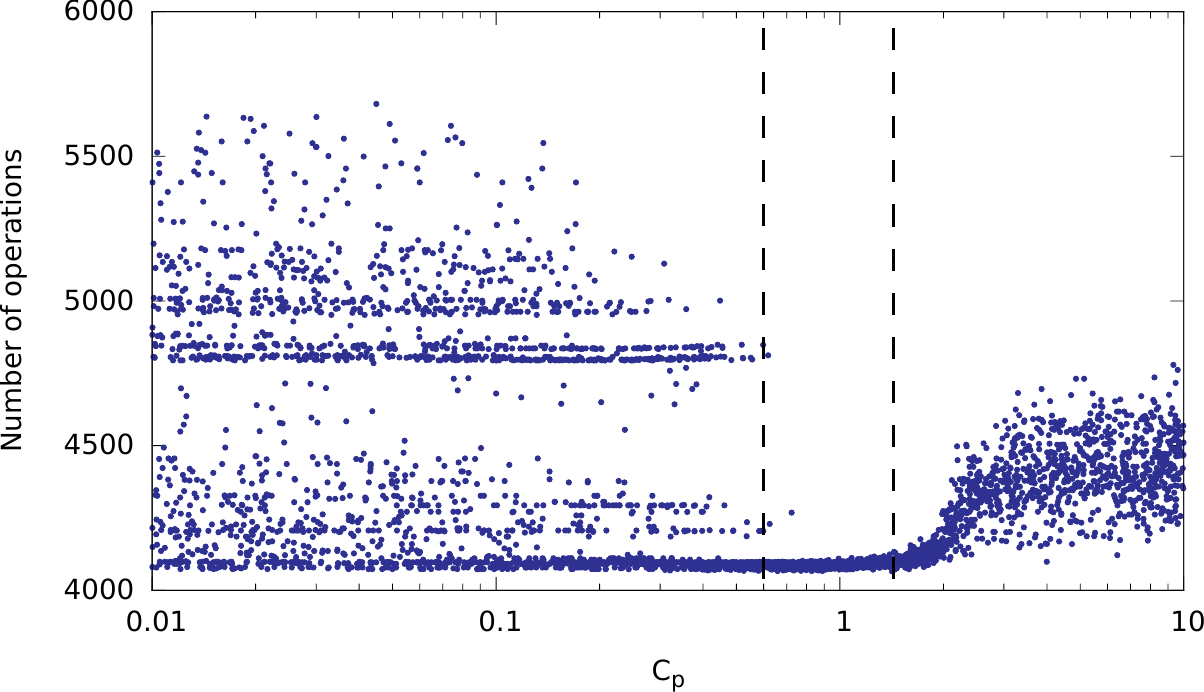} }
\caption{A sensitivity analysis for the expression HEP($\sigma$) of the exploration-exploitation constant $C_p$ horizontally, on the number of operations (lower is better) vertically. The figure on the left has 300 iterations and the right figure has 3000 iterations. Each figure displays 4000 MCTS runs (dots). There are three regions: at low $C_p$ there is a band structure of local minima, at high $C_p$ there is a diffuse region, and, if the number of iterations is high enough, at an intermediate $C_p$ there is a region where the probability of finding the global minimum is high. This region is demarcated by dashed lines in the picture on the right.}
\label{fig:sigma_scatter}
\end{figure}

Below, we will investigate the effect of (1) different values of $C_p$ and (2) the number of iterations $N$ on the number of operations by performing a sensitivity analysis. A polynomial from high energy physics called HEP($\sigma$) is taken as our benchmark \cite{Kuipers2013}. In figure \ref{fig:sigma_scatter} the result is displayed for $N=300$ on the left and $N=3000$ on the right. On the x-axis we indicate the $C_p$ values ranging from $0.01$ to $10$; on the y-axis we indicate the number of operations, ranging from 4000 to 6000 (lower is better). We see that for low $C_p$ there are many local minima, as evidenced by the clustering in horizontal lines (thick clusters and sparse clusters). If the $C_p$ is high, there is much exploration and this translates to a diffuse region where there is no convergence to a local minimum. When the $N$ is sufficiently high (e.g., $N=3000$), we see an intermediate region emerging where the probability of finding the global minimum is high (only one thick line). This region is demarcated by dotted lines in figure \ref{fig:sigma_scatter}.

\begin{figure}
\centering
\includegraphics[scale=1.0]{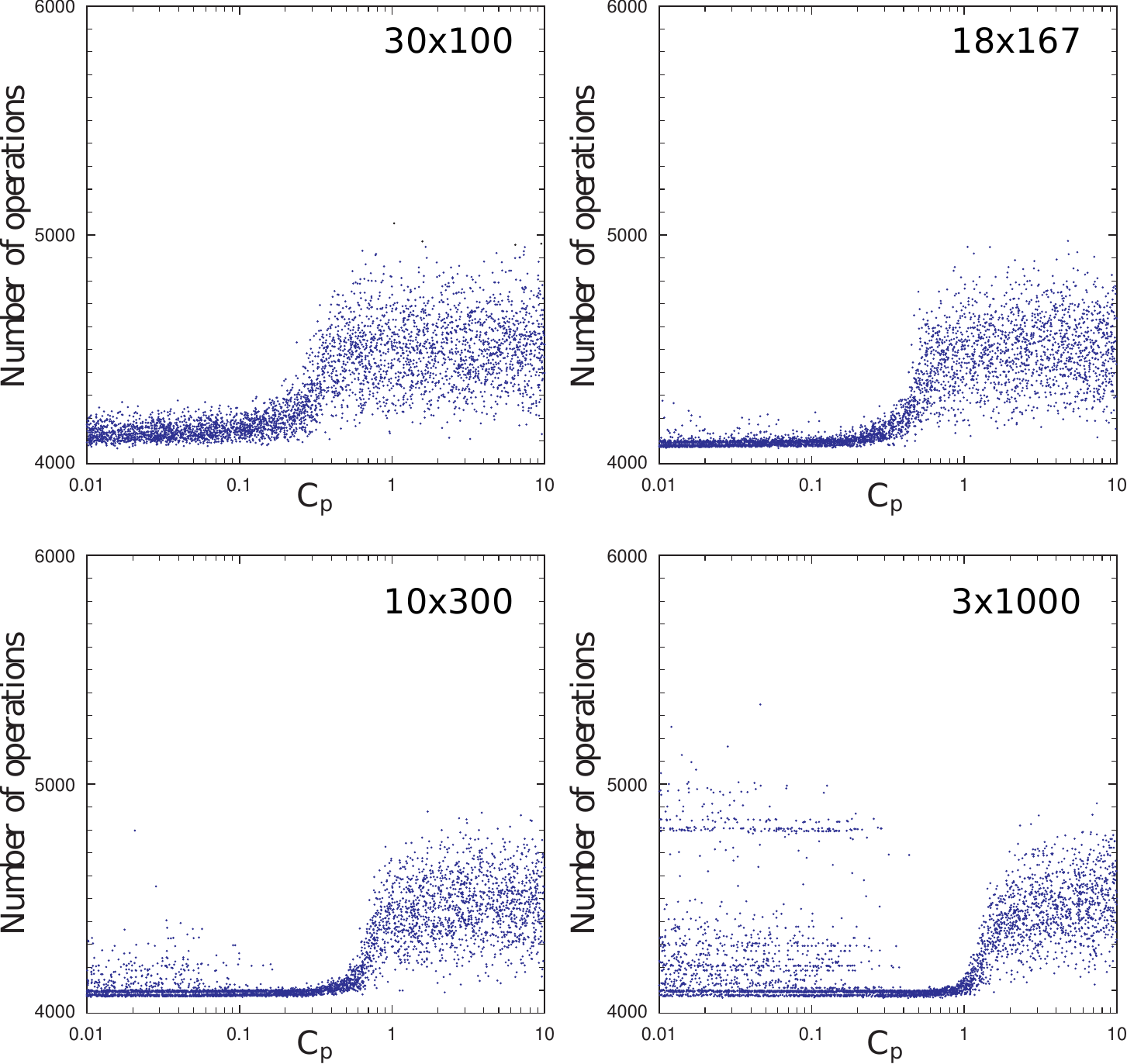}
\caption{A sensitivity analysis for HEP($\sigma$) of the exploration-exploitation constant $C_p$ horizontaly, on the number of operations (lower is better) vertically. Each dot is the best score of $R$ MCTS runs with each $N$ number of iterations. The total number of iterations per dot $R \times N$ is $3000$ for all graphs. The top left figure has 30 repetitions and 100 expansions, the top right one has 18 repetitions and 167 expansions, the bottom left one has 10 repetitions and 300 expansions and the bottom right figure has 3 repetitions and 1000 expansions.}
\label{fig:sigma_rep}
\end{figure}

From figure \ref{fig:sigma_scatter} we see that if $C_p$ is in the low or intermediate region, the density of dots is highest near the global minimum. This means that if we select the best result of several simulations, the probability of finding the global minimum is high. To study these effects, we have made scatter plots of the number of operations and $C_p$, given a number of repetitions $R$ and a number of iterations $N$, while keeping the total number of iterations $R \times N$ constant \cite{Herik2013B}. The best value of these $R \times N$ runs is selected and is shown in figure \ref{fig:sigma_rep}, for HEP($\sigma$). On the top left, we have 30 runs with 100 expansions ($30 \times 100$), on the top right 18 runs of 167 expansions ($18 \times 167$), on the bottom left 10 runs of 300 expansions ($10 \times 300$) and on the bottom right 3 runs of 1000 expansions ($3 \times 1000$). Each graph contains 4000 measurements (dots) with 3000 iterations in total for each measurement. Thus, this graph is comparable in CPU time to MCTS with 3000 iterations ($1 \times 3000$), displayed in figure \ref{fig:sigma_scatter} on the right. We notice that for all but the bottom right graph, the local minima have almost disappeared for low $C_p$, so that only a cluster around the global minimum remains. For the bottom right graph the local minima are very sparse. The top right graph and the bottom left graph have the best characteristics: from $C_p=0.01$ to $C_p=0.3$ the probability of finding the local minimum is higher than the MCTS run with $3000$ iterations. The main obstacle is that we do not know in advance how many repetitions or which number of iterations we should use, without computing the scatter plots.

\begin{table*}[t]
\centering
\begin{tabular}{|l|c|c||c||c|c|}
\cline{2-6}
\multicolumn{1}{c|}{} & \# vars & original & occurrence & MCTS 1k & MCTS 10k\\
\hline
res(7,4)    & 13  &   29\,163   & 4\,968 & $(3.86 \pm 0.1)\cdot 10^3$ & $(3.84 \pm 0.01)\cdot 10^4$\\
res(7,5)    & 14  &  142\,711    & 20\,210  & $(1.39 \pm 0.01)\cdot 10^4$ & $13768 \pm 28$\\
res(7,6)    & 15  & 587\,880    & 71\,262   & $(4.58 \pm 0.05)\cdot10^4$ & $(4.54 \pm 0.01)\cdot10^4$\\
\hline
HEP($\sigma$) & 15 & 47\,424  & 6744  & $4114 \pm 14$ & $4 087 \pm 5$\\
HEP($F_{13}$) & 24 & 1\,068\,153 & 92\,617  & $(6.6 \pm 0.2)\cdot 10^4$ & $(6.47\pm 0.08)\cdot 10^4$\\
HEP($F_{24}$) & 31 & 7\,722\,027 & 401\,530  & $(3.80 \pm 0.06)\cdot 10^5$ & $(3.19 \pm 0.04)\cdot 10^5$\\
\hline
\end{tabular}

\caption{Results of MCTS with 1000 iterations and 10\,000 iterations compared to occurrence order and the original number of operations. The MCTS numbers are statistical averages and standard deviations.}
\label{tbl:perf}
\end{table*}

In table \ref{tbl:perf} the results (from \cite{Kuipers2013}) are shown for MCTS runs on several polynomials. The results are statistical averages. The res(m, n) polynomials are resolvents and are defined by $\text{res}(m, n) = \text{res}_x(\sum^m_{i=0} a_ix^i, \sum^n_{i=0} b_ix^i)$, as described in \cite{Leiserson2010}. The HEP polynomials stem from theoretical predictions of scattering processes in high energy physics. The $C_p$ used in the above results has been manually tuned to the region where good values are obtained. Additionally, the construction direction of the scheme was selected appropriately (see section \ref{sec:issues}).

We see that MCTS with 10\,000 iterations reduces the number of operations by 2400\% compared to the original and reduces it 25\% more than the occurrence order scheme for HEP($F_{24}$). For the resolvent res(7,6) the reduction is 1300\% compared to the original and 56\% compared to the occurrence order scheme. In practice, numerical integration of these expressions could take weeks, therefore the simplifications are able to reduce computation time from weeks to days.

\subsection{Unresolved Issues}
\label{sec:issues} 

There are two issues with the current form of the MCTS algorithm. First of all, the $C_p$ parameter must be tuned. Sometimes the region of $C_p$ that yields good values is small, thus it may be computationally expensive to find an appropriate $C_p$. Second, trees are naturally asymmetric: there is more exploration at nodes close to the root compared to the nodes deeper in the tree. Moreover, only a few branches are fully expanded to the bottom. Consequently, the final variables in the scheme will be filled out with the variables of a random playout. No optimization is done at the end of these branches. As a result, if a very specific order of moves at the end of the tree yields optimal results, this will not be found by MCTS. The issue can be partially reduced by adding a new parameter that specifies whether the Horner scheme should be constructed in reverse, so that the variables selected near the root of the tree are actually the last to be extracted \cite{Kuipers2013, Ruijl2014}.

\begin{figure}[ht]
\centering
\subfigure[Forward scheme]{ \includegraphics[scale=0.572]{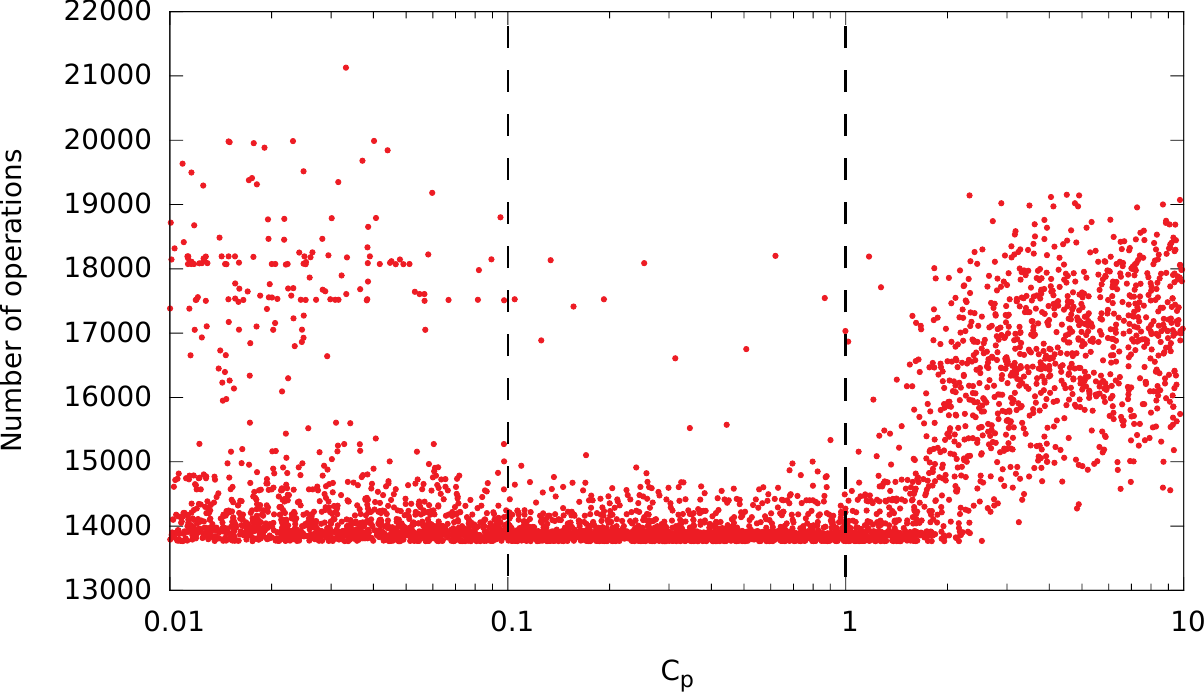} \label{fig:backa}}
\subfigure[Backward scheme]{ \includegraphics[scale=0.572]{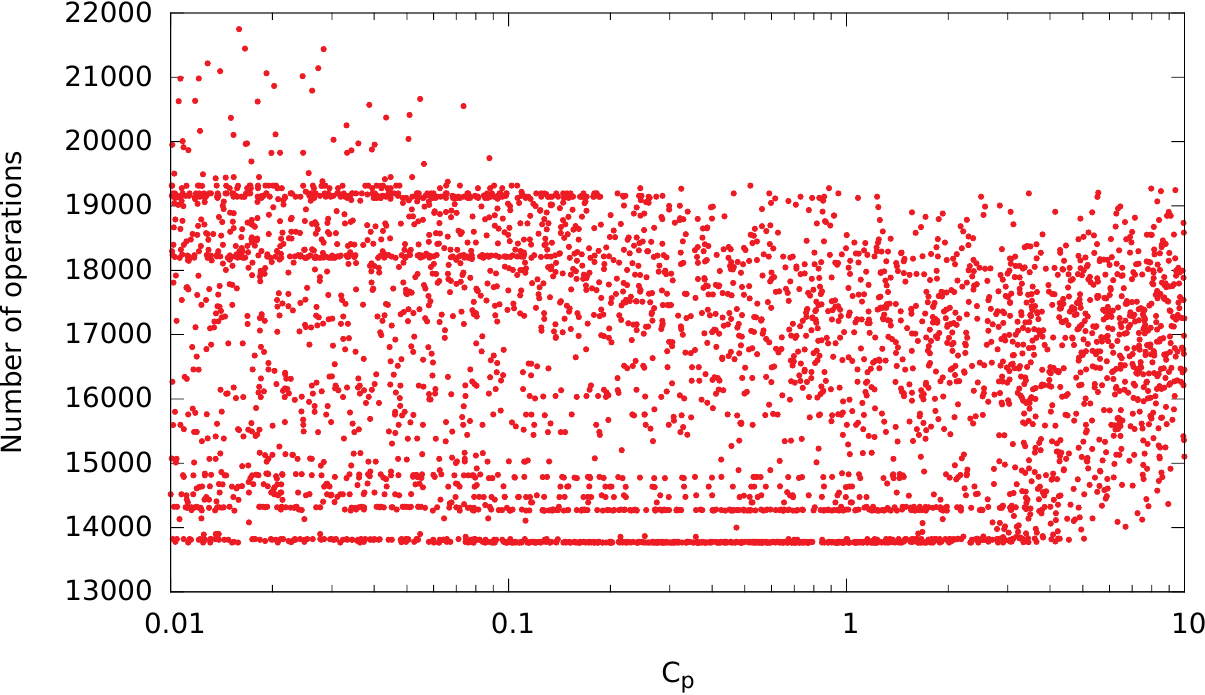} \label{fig:backb}}
\caption{res(7,5): differences between forward (left) and backward (right) Horner schemes, at $N=1000$ tree updates with SA-UCT. Forward Horner schemes generate a region of $C_p$ where the number of operations is near the global minimum, whereas backward schemes have multiple high-density local minima and a diffuse region.}
\label{fig:fwd_vs_bkwd}
\end{figure}

In figure \ref{fig:fwd_vs_bkwd} the difference between a forward and a backward MCTS search with $1000$ updates is shown for the polynomial res(7,5) in scatter plot. For the forward construction, we see that there is a region in $C_p$ where the results are good: the optimum is found often. However, the backward scheme does not have a similar range. For other polynomials, it may be better to use the backward scheme, as is the case for HEP($\sigma$) and HEP($F_{13}$). Currently, there is no known way to predict whether forward or backward construction should be used. Thus, this introduces an extra parameter to our algorithm.

Even though the scheme direction parameter reduces the problem somewhat, the underlying problem that there is little exploration at the end of the tree still remains. To overcome the issues of tuning $C_p$ and the lack of exploration, we have looked at a related tree search method called Nested Monte Carlo Search.

\section{Nested Monte Carlo Search}
\label{sec:nmcs}
Nested Monte Carlo Search (NMCS) addresses the issue of the exploration bias towards the top of the tree by sampling all children at every level of the tree \cite{Cazenave2009}. In its simplest form, called a level-1 search, a random playout is performed for each child of a node. Next, the child with the best score is selected, and the process is repeated until one of the end states is reached. This method can be generalized to a level $k$ search, where the above process is nested: a level $k$ search chooses the best node from a level $k-1$ search performed on its children. Thus, if the NMCS level is increased, the top of the tree is optimized with greater detail. Even though NMCS makes use of random playouts, it does so at every depth of the tree as the search progresses. Consequently, there is always exploration near the end of the tree.

\begin{figure}
\centering
\includegraphics[scale=0.5]{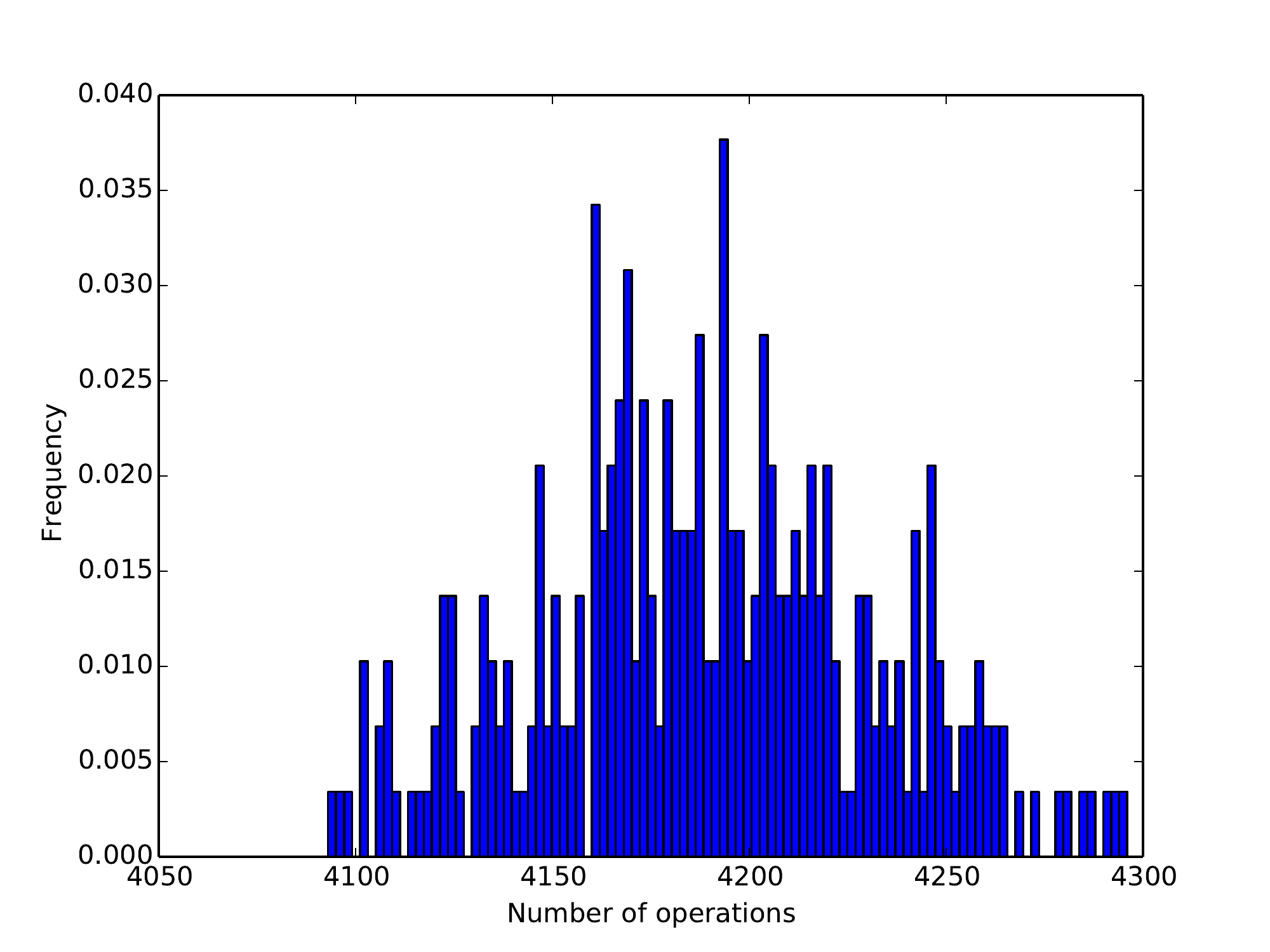}
\caption{NMCS level 2 for HEP($\sigma$), taking 8500 evaluations. This is comparable in CPU time to MCTS with 8500 runs. The number of operations is $4189 \pm 43$, averaged over 292 samples.}
\label{fig:sigma_nmcs}
\end{figure}

In figure \ref{fig:sigma_nmcs} the results for level 2 NMCS are shown for HEP($\sigma$). The average number of operations is $4189 \pm 43$. To compare the performance of NMCS to that of MCTS, we study the run-time. Since most of the run-time is spent on the evaluation function, we may compare the number of evaluations instead. A level-2 search for HEP($\sigma$) takes $8500$ evaluations. In order to be on a par with MCTS, the score should have been between MCTS $1000$ and MCTS $10\,000$ iterations. However, we see that the score is higher than MCTS with $1000$ iterations and thus we may conclude that the performance of NMCS is inferior to MCTS for HEP($\sigma$). 

We have performed similar experiments with NMCS on other polynomials, but the resulting average number of operations were always greater than MCTS's. The reason is likely because we select a low level $k$: a level-1 search selects the best child using one sample per child, a process which is highly influenced by chance. However, there are some performance issues with using a higher $k$. To analyze these, we investigate the number of evaluations that a level $n$ search requires.

The form of our tree is known, since every variable should appear only once. This means that there are $n$ children at the root, $n-1$ children of children, and $n-d$ children at depth $d$. Thus a level-1 search takes $n + (n-1) + (n-2) + \ldots +1 = n (n + 1)/2$ evaluations. It can be shown that a level $k$ search takes $S_n^{k + n}$, where $S$ is the Stirling Number of the First Kind. This number grows rapidly: if $k=1$ and $n=15$, the number of evaluations is $120$, and for level $k=2$, it takes $8500$ evaluations. For an expression with $100$ variables, a level-1 search takes $5050$ evaluations, and a level-2 search takes $13\,092\,125$ evaluations. 

The evaluation function is expensive for our polynomials: HEP($F_{13}$) takes about 1 second per evaluation and HEP($F_{24}$) takes $6.6$ seconds. We have experimented with parallelizing the evaluation function, but due to the fine-grained nature of the evaluation function, this was unsuccessful. For HEP($F_{24}$) a million iterations will be slow, hence for practical reasons we have only experimented with a level-1 search. 

The domains in which NMCS performed well, such as Morpion Solitaire and SameGame, have a cheap evaluation function relative to the tree construction \cite{Cazenave2009}. If the evaluation function is expensive, even the level-1 search takes a long time. 

Based on the remarks above, we may conclude that for polynomials with a large number of variables, NMCS becomes unfeasible. 

\section{SA-UCT}
\label{sec:sa-uct}
Since NMCS is unsuitable for simplifying large expressions, we return our focus to MCTS, but this time on the UCT best child criterion. We now consider the role of the exploration-exploitation constant $C_p$. We notice that in the beginning of the simulation there is as much exploration as there is at the end, since $C_p$ remains constant throughout the search. For example, the final 100 iterations of a 1000 iterations MCTS run are used to explore new branches even though we know in advance that there is likely not enough time to reach the final nodes. Thus we would like to modify the $C_p$ to change during the simulation to emphasize exploration early in the search and emphasize exploitation towards the end.

We introduce a new, dynamic exploration-exploitation parameter $T$ that decreases linearly with the iteration number:
\begin{equation}
T(i) =  C_p \frac{N - i}{N}
\label{eq:t}
\end{equation}
where $i$ is the current iteration number, $N$ the preset maximum number of iterations, and $C_p$ the initial exploration-exploitation constant at $i=0$. 

We modify the UCT formula to become:
\begin{equation}
\underset{\text{children $c$ of $s$}}{\operatorname{argmax}} \bar{x}(c) + 2 T(i) \sqrt{\frac{2 \ln n(s)}{n(c)}}
\label{eq:sa-uct}
\end{equation}
where $c$ is a child of node $s$, $\bar{x}(c)$ is the average score of child $c$, $n(c)$ the number of visits at node $c$, and $T(i)$ the dynamic exploration-exploitation parameter of formula (\ref{eq:t}).

The role of $T$ is similar to the role of the temperature in Simulated Annealing: in the beginning of the simulation there is much emphasis on exploration, the analogue of allowing transitions to energetically unfavourable states. During the simulation the focus gradually shifts to exploitation, analogous to annealing. Hence, we call our new UCT formula ``Simulated Annealing UCT (SA-UCT)''.

We notice four improvements over UCT: not only are the final iterations used effectively, there is more exploration in the middle and at the bottom of the tree. This is due to more nodes being expanded at lower levels, because the $T$ is lowered. As a consequence, we see that more branches reach the end states. As a result, there is exploration near the bottom, where there was none for the random playouts.

\begin{figure*}[ph]
\centering
\textbf{res(7,5) with 14 variables} \par\medskip
\subfigure[$N=300$]{ \includegraphics[scale=0.572]{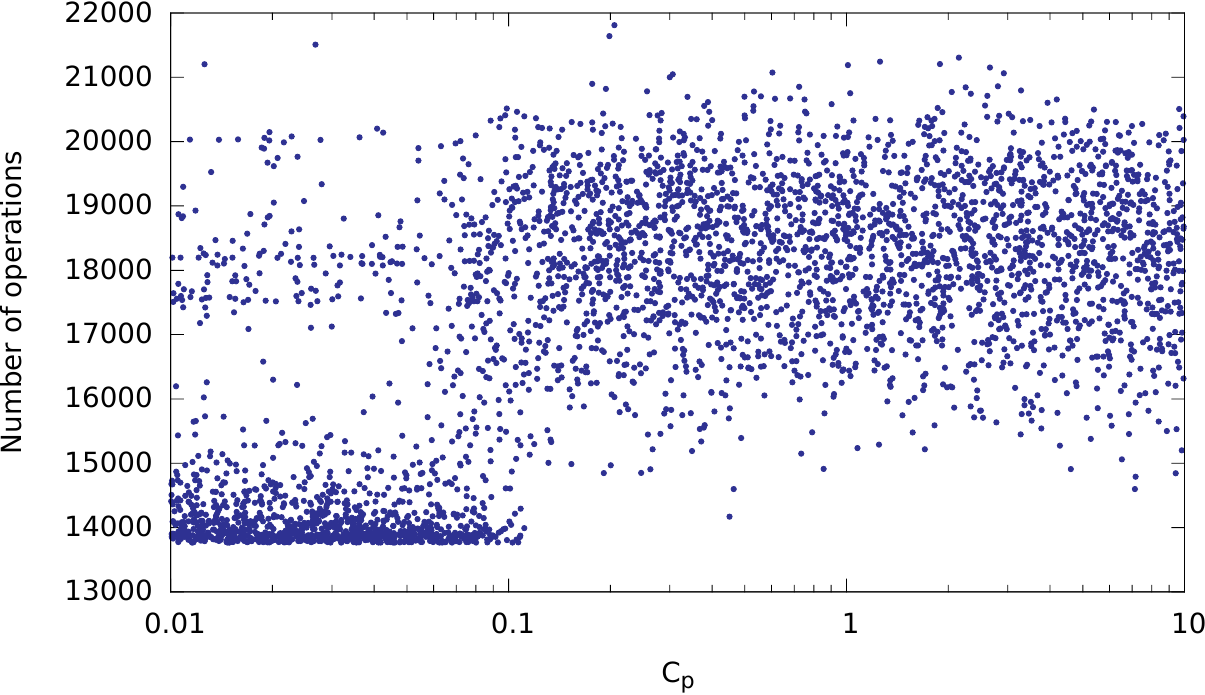} }
\subfigure[$N=300$]{ \includegraphics[scale=0.572]{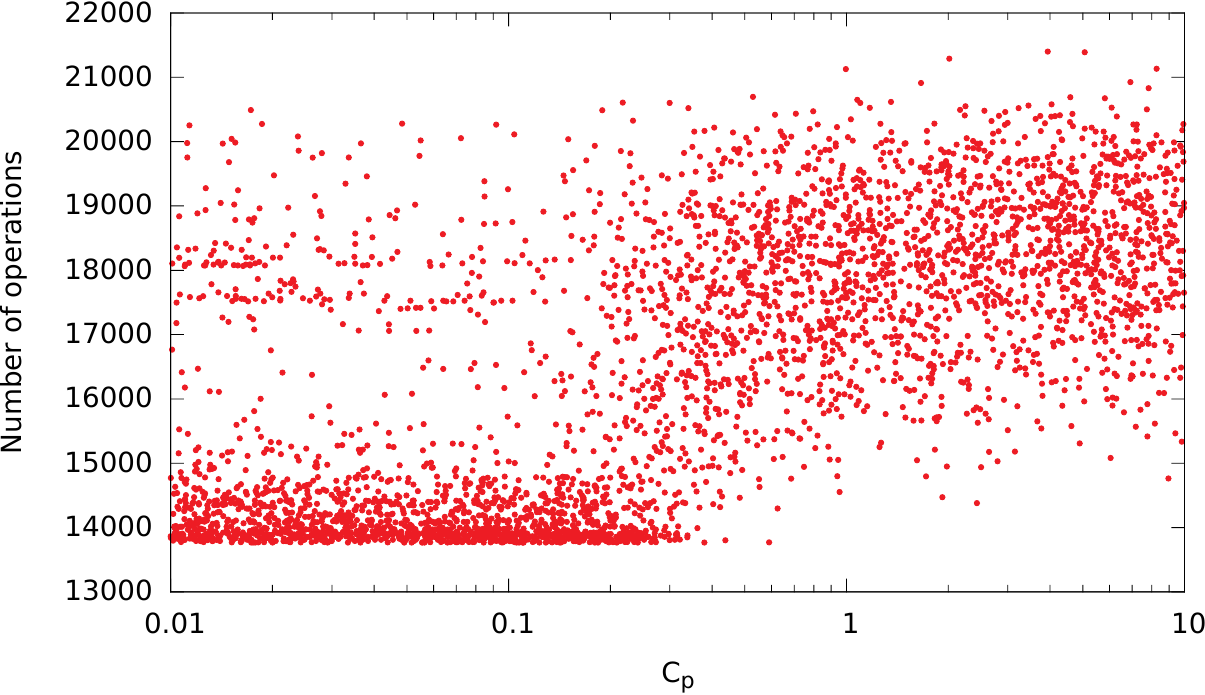} }\\ \bigskip
\subfigure[$N=1000$]{ \includegraphics[scale=0.572]{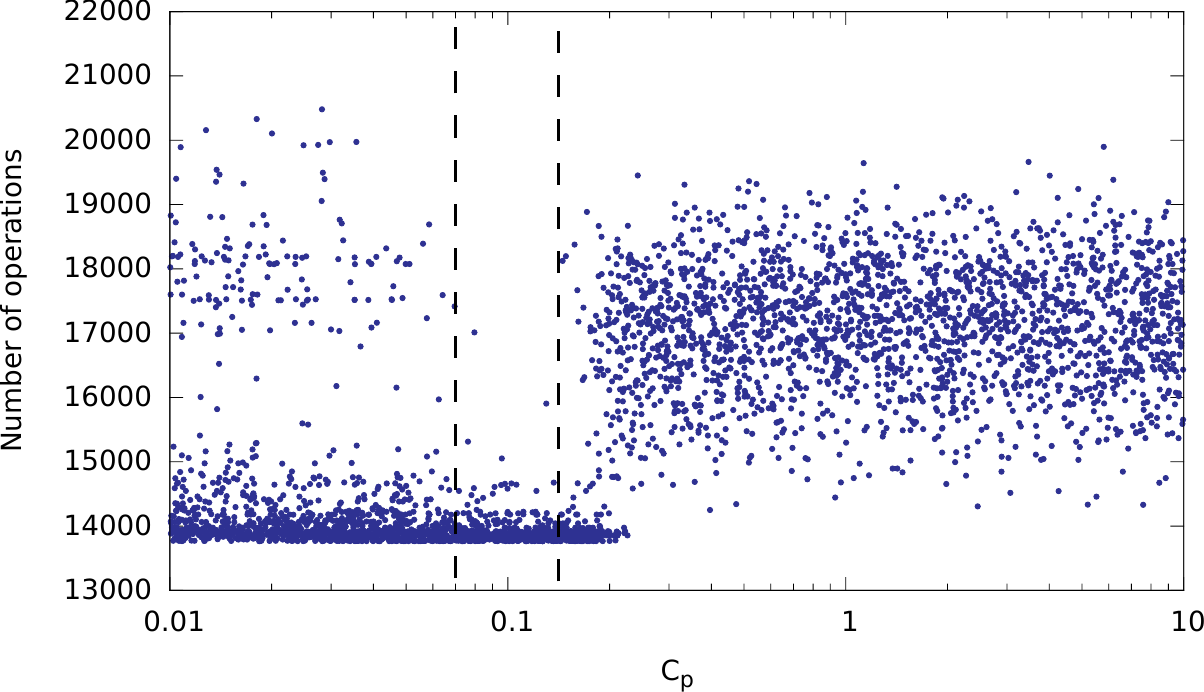} }
\subfigure[$N=1000$]{ \includegraphics[scale=0.572]{data/res75_1000_fwd.pdf} }\\ \bigskip
\subfigure[$N=3000$]{ \includegraphics[scale=0.572]{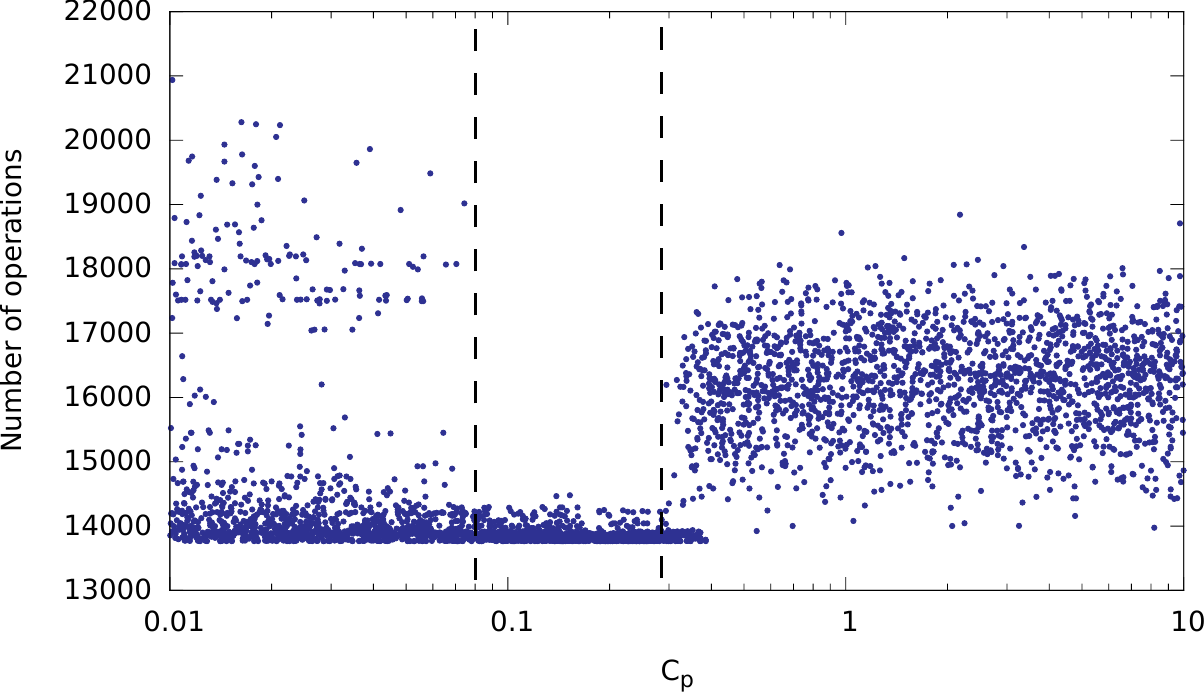} }
\subfigure[$N=3000$]{ \includegraphics[scale=0.572]{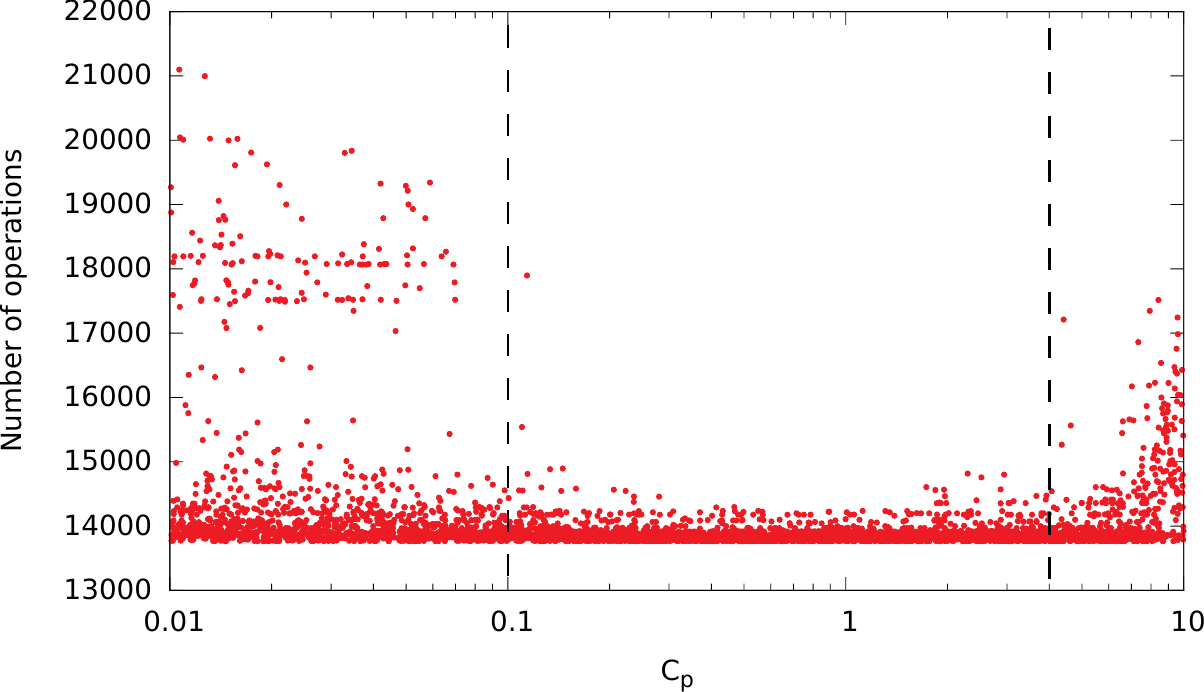} }\\
\caption{res(7,5) polynomial with 14 variables: on the x-axis we show $C_p$ and on the y-axis the number of operations. A lower number of operations is better. On the left, we show UCT with constant $C_p$ and on the right we show SA-UCT where $C_p$ is the starting value of $T$. Each graph contains 4000 runs (dots) of MCTS. Figure \ref{fig:res75}(a) and \ref{fig:res75}(b) are measured with $N=300$ tree updates, \ref{fig:res75}(c) and \ref{fig:res75}(d) with $N=1000$, and \ref{fig:res75}(e) and \ref{fig:res75}(f) with $N=3000$ updates. As indicated by the dashed lines, an area with an operation count close to the global minimum appears, as soon as there are sufficient tree updates $N$. This area is wider for SA-UCT than for UCT.}
\label{fig:res75}
\end{figure*}

In order to analyze the effect of SA-UCT on the fine-tuning of $C_p$ (the initial temperature), we perform a sensitivity analysis on $C_p$ and $N$ \cite{Ruijl2014}. In figure \ref{fig:res75} the results for the res(7,5) polynomial with 14 variables are displayed. Horizontally, we have $C_p$, and vertically we have the number of operations (where less is better). A total of 4000 MCTS runs (dots) are performed for a $C_p$ between $0.001$ and $10$. On the left we show the results for UCT and on the right for SA-UCT. Just as in figure \ref{fig:sigma_scatter}, we identify a region with local minima for low $C_p$, a diffuse region for high $C_p$ and an intermediate region in $C_p$ where good results are obtained. This region becomes wider if the number of iterations $N$ increases, for both SA-UCT and UCT.  

However, we notice that the intermediate region is wider for SA-UCT, compared to UCT. For $N=1000$, the region is $[0.1,1.0]$ for SA-UCT, whereas it is $[0.07,0.15]$ for UCT. Thus, SA-UCT makes the region of interest about $11$ times larger for res(7,5). This stretching is not just an overall rescaling of $C_p$: the uninteresting region of low $C_p$ did not grow significantly. For $N=3000$, the difference in width of the region of interest is even larger.

\begin{figure*}[ph]
\centering
\textbf{HEP($\sigma$) with 15 variables} \par\medskip
\subfigure[$N=300$]{\includegraphics[scale=0.572]{data/sigma_300_bkwd_cons.pdf}}
\subfigure[$N=300$]{ \includegraphics[scale=0.572]{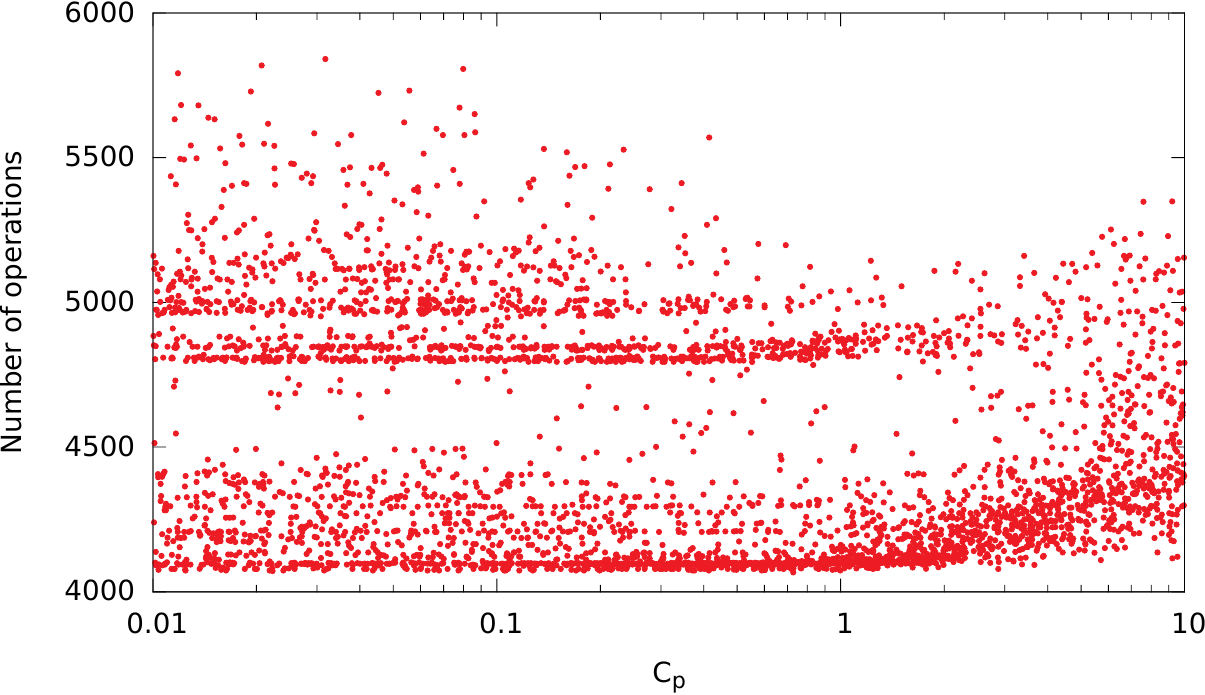}}\\ \bigskip
\subfigure[$N=1000$]{ \includegraphics[scale=0.572]{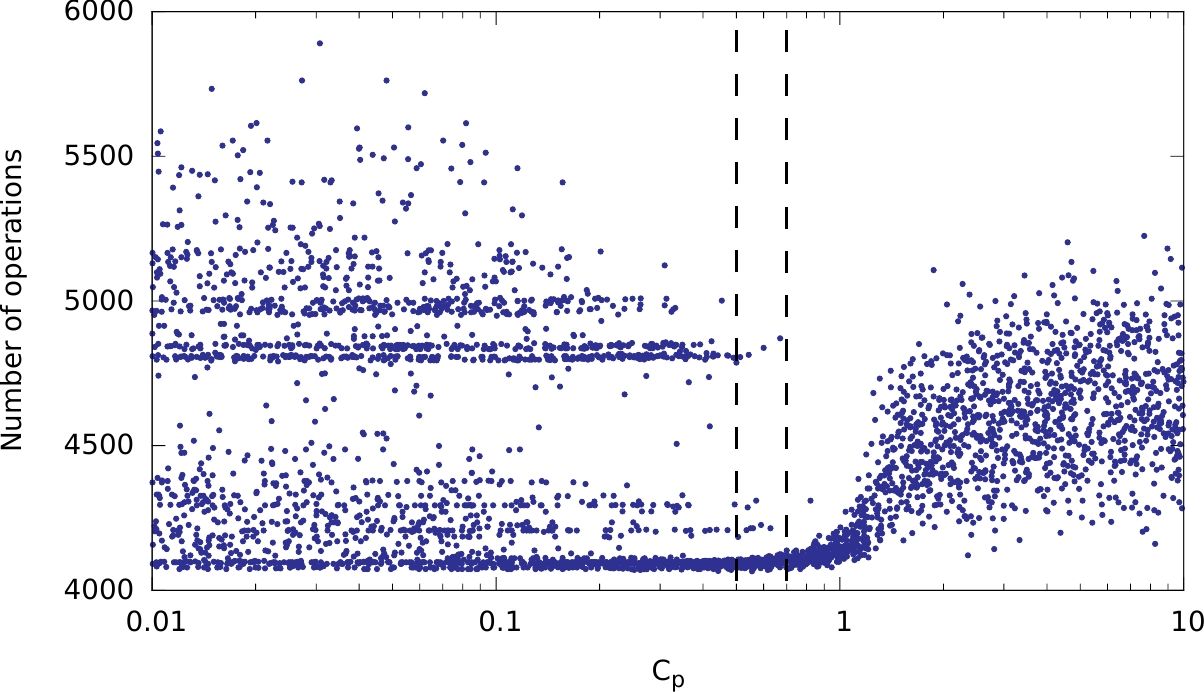}}
\subfigure[$N=1000$]{ \includegraphics[scale=0.572]{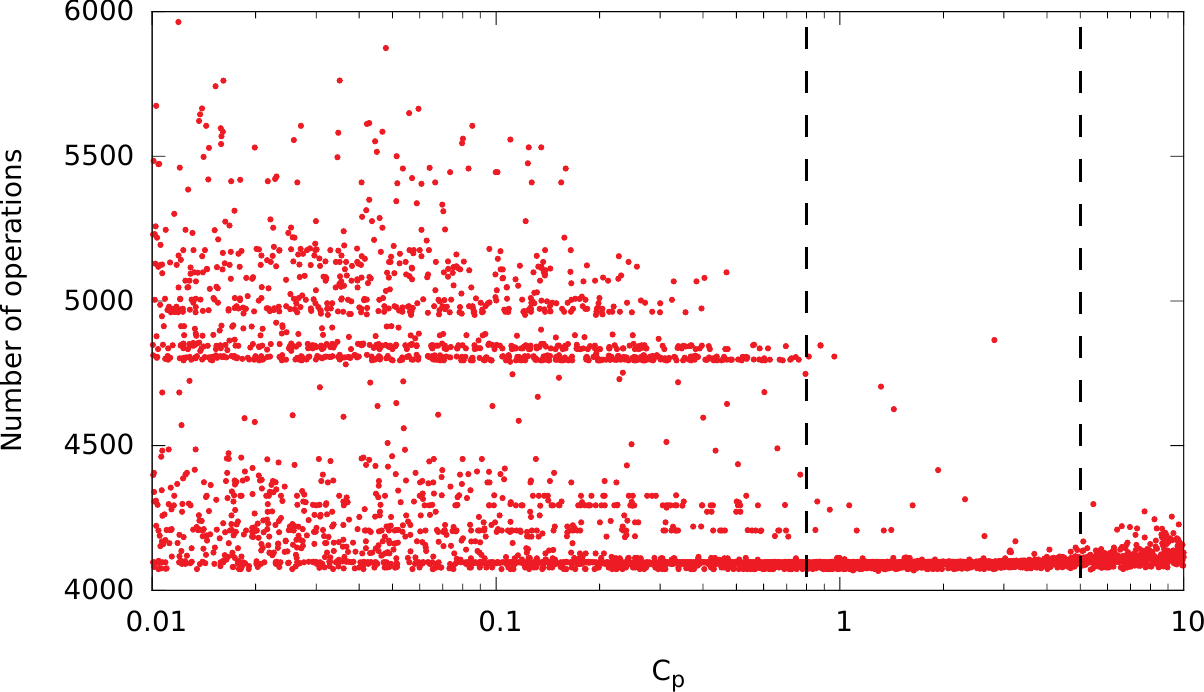}}\\ \bigskip
\subfigure[$N=3000$]{ \includegraphics[scale=0.572]{data/sigma_3000_bkwd_cons.pdf}}
\subfigure[$N=3000$]{ \includegraphics[scale=0.572]{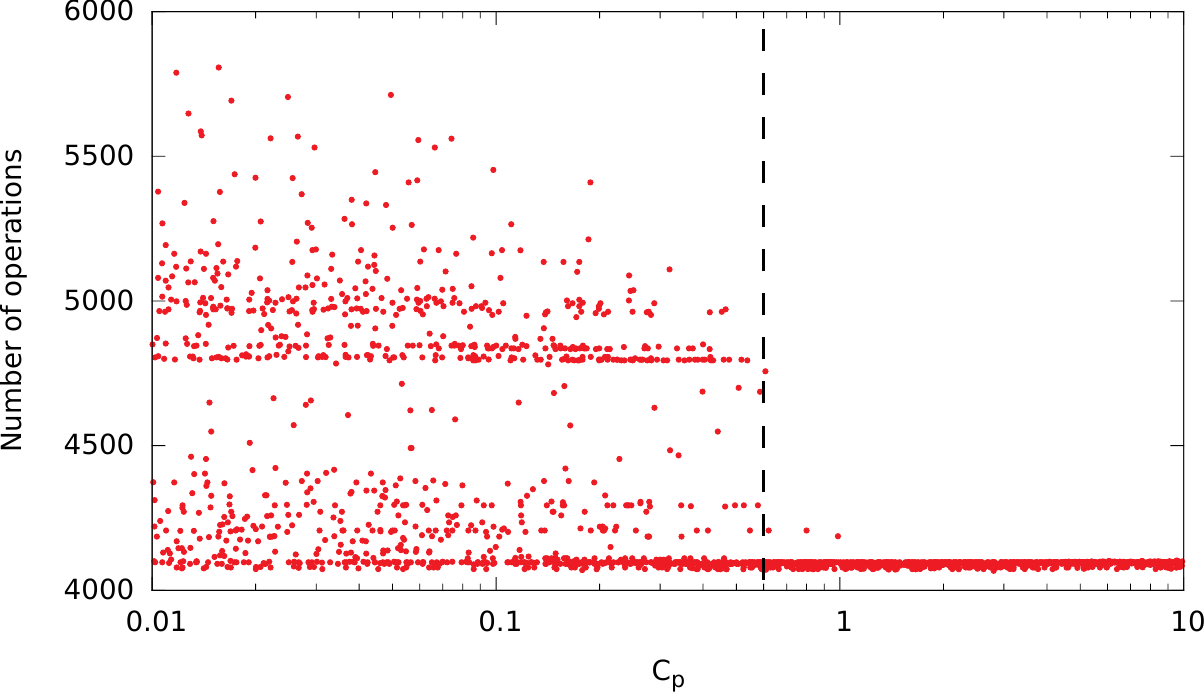}}\\

\caption{HEP($\sigma$) with 15 variables: on the x-axis we show $C_p$ and on the y-axis the number of operations. A lower number of operations is better. On the left, we show UCT with constant $C_p$ and on the right we show SA-UCT where $C_p$ is the starting value of $T$. Each graph contains 4000 runs (dots) of MCTS. Figure \ref{fig:sigma}(a) and \ref{fig:sigma}(b) are measured with $N=300$ tree updates, \ref{fig:sigma}(c) and \ref{fig:sigma}(d) with $N=1000$, and \ref{fig:sigma}(e) and \ref{fig:sigma}(f) with $N=3000$ updates. As indicated by the dashed lines, an area with an operation count close to the global minimum appears, as soon as there are sufficient tree updates $N$. This area is wider for SA-UCT than for UCT.}
\label{fig:sigma}
\end{figure*}

In figure \ref{fig:sigma}, we show a similar sensitivity analysis for HEP($\sigma$) with 15 variables. We identify the same three regions and see that the region of interest is $[0.5,0.7]$ for UCT and $[0.8,5.0]$ for SA-UCT at $N=1000$. This means that the region is about $20$ times larger relative to the uninteresting region of low $C_p$, which grew from $0.5$ to $0.8$. We have performed the experiment on more than five other expressions, and we obtain similar results \cite{Ruijl2014}.

\newpage

\section{Conclusion}
\label{sec:conclusion}
MCTS is able to find Horner schemes that yield a smaller number of operations than the naive occurrence order schemes. For some polynomials, MTCS yields reductions of more than a factor 24. However, this method has two issues that deserve closer inspection, viz. fine-tuning the exploration-exploitation constant $C_p$ and the fact that there is little exploration at the bottom of the tree.

We attempted to address these issues by using NMCS, but found that this method is unsuitable for our domain due to a slow evaluation function.

Next, we modified $C_p$ to decrease linearly with the iteration number. We call this new selection criterion SA-UCT. SA-UCT caused more branches to reach end states and simplified the tuning of $C_p$: the region of appropriate $C_p$ was increased by at least a tenfold.

As a conclusion we may state that using SA-UCT, we are able to simplify expressions by at least a factor of 24 for our real-world test problems, with a reduced sensitivity to $C_p$, effectively making numerical integration 24 times faster. 

\section{Discussion / Future Work}
\label{sec:discussion}
Using SA-UCT, the fine-tuning of the exploration-exploitation constant $C_p$ has become easier, but still there is no automatic way of tuning it. A straightforward binary search may be suitable, if the region of $C_p$ where good results are obtained is large enough. Moreover, it is still an unresolved question whether a forward or backward construction has to be chosen. Perhaps the preferred scheme orientation can be derived from some currently unidentified structure of the polynomial.

Furthermore, to come to a deep insight of the choices to be made, our methods have to be tested on more expressions from various field. Currently, we have expressions from high energy physics and the resolvents from mathematics, but our methods could also be tried on (1) a class of boolean expressions, and (2) on results from astrophysics.


Additional work has to be spent on supporting non-commutative variables and functions. In principle, Horner schemes can be applied to expressions with non-commuting variables if the Horner scheme itself only consists of commuting variables. Furthermore, the common subexpression elimination should take into account that some variables do not commute. For example, in figure \ref{fig:cse}, the highlighted part is not a common subexpression if $b$ and $c$ are non-commutative. 

Next, it is challenging to investigate replacing variables by a linear combination of other variables. These global patterns are not recognized by CSEE and may reduce the number of operations considerably. However, determining which variables to combine linearly, might be time consuming.

In the future, project HEPGAME will focus on solving a range of different problems in high energy physics \cite{hepgame}. Polynomial reduction is a first step, but there are many other challenges, such as solving recurrence relations with a minimal number of generated terms. The solution to this class of problem allows for previously impossible computations of the integrals involved in higher order correction terms of scattering processes at CERN.

\section{Acknowledgements}
This work is supported in part by the ERC Advanced Grant no. 320651, ``HEPGAME''.

\bibliographystyle{abbrv}
{\small
\bibliography{ref}}

\end{document}